\documentclass[sigconf]{acmart}
\usepackage{subfig}
\usepackage{subcaption}
\usepackage{multirow}
\AtBeginDocument{%
  }

\setcopyright{acmlicensed}
\copyrightyear{2026}
\acmYear{2026}
\acmDOI{XXXXXXX.XXXXXXX}
\acmConference[Conference BIOKDD 2026]{25th International Workshop on Data Mining in Bioinformatics in conjunction of SIGKDD 2026}{Aug 09--13,
  2026}{Jeju, Korea}
\acmISBN{978-1-4503-XXXX-X/2018/06}

\begin{document}

\title{A Temporal Machine Learning-Based Time-to-Event Model for Predicting ALS Progression and Healthcare Utilization}

\author{Zongliang Yue$^{\dag}$,$^{*}$}

\affiliation{%
  \institution{Department of Health Outcomes Research and Policy, Harrison College of Pharmacy, Auburn University}
  \city{Auburn}
  \state{AL}
  \country{USA}
}
\email{zzy0065@auburn.edu}

\author{Qi Li$^{\dag}$}
\affiliation{%
  \institution{Mathematics and Computer Science Department, Fisk University}
  \city{Nashville}
  \state{TN}
  \country{USA}
}

\author{Terry Heiman-Patterson}
\affiliation{%
  \institution{Lewis Katz School of Medicine, Temple University}
  \city{Philadelphia}
  \state{PA}
  \country{USA}
}

\author{Frank Bearoff}
\affiliation{%
  \institution{Lewis Katz School of Medicine, Temple University}
  \city{Philadelphia}
  \state{PA}
  \country{USA}
}

\author{Zhaohui Qin}
\affiliation{%
  \institution{Department of Biostatistics and Bioinformatics, Emory University}
  \city{Atlanta}
  \state{GA}
  \country{USA}
}

\author{Huanmei Wu$^{*}$}
\affiliation{%
  \institution{Barnett College of Public Health, Temple University}
  \city{Philadelphia}
  \state{PA}
  \country{USA}
}
\email{huanmei.wu@temple.edu}

\renewcommand{\shortauthors}{Yue et al.}

\begin{abstract}
\footnote{$\dag$: equal contribution; $*$: corresponding author.}
Amyotrophic lateral sclerosis (ALS) is a progressive and heterogeneous neurodegenerative disease in which predicting clinically meaningful milestones, such as assistive device use, remains challenging. We developed a time-to-event, digital-twin-inspired framework that integrates longitudinal ALS Functional Rating Scale–Revised (ALSFRS-R) trajectories with survival modeling to support individualized prediction of functional decline and assistive device utilization. We constructed a harmonized longitudinal dataset by integrating diagnosis records, ALSFRS-R assessments, activities of daily living, and demographic information, followed by preprocessing to ensure data quality, temporal alignment, and cohort consistency. Correlation-based clustering identified coherent functional domains spanning bulbar, upper limb, axial, lower limb, and respiratory systems. Generalized additive mixed models characterized nonlinear, domain-specific functional decline across all domains. In addition, a temporal machine learning model was developed to predict longitudinal functional decline and capture stage-dependent disease progression. Cox proportional hazards modeling further identified lower limb function, particularly walking and stair climbing, as the strongest predictors of earlier wheelchair access. Building on these results, we implemented a digital twin-inspired temporal machine learning-based time-to-event (TTE) model that generates individualized survival curves and dynamically predicts wheelchair-free survival (http://als-digital-twin-app.streamlit.app). This framework provides a scalable, interpretable, and clinically actionable approach for linking ALS progression with personalized decision support, with applications in proactive care planning, clinical trial stratification, and precision medicine.
\end{abstract}


\maketitle
\section{Introduction}

\label{sec:intro}

Amyotrophic lateral sclerosis (ALS) is a progressive neurodegenerative disorder characterized by the degeneration of upper and lower motor neurons, leading to muscle weakness, loss of mobility, respiratory failure, and ultimately death~\cite{feldman2022als,kiernan2011als,vanes2017als}.
The disease affects approximately 2.2 individuals per 100,000 in Europe
and 1.9 per 100,000 in the United States annually and remains without a
curative therapy.
The median survival ranges from 2 to 5 years following
diagnosis~\cite{marcu2024diagnostic}.
Clinically, ALS exhibits substantial heterogeneity in both presentation
and progression, including variability in site of onset (bulbar vs.\
limb), rate of functional decline, and patterns of system involvement
across motor and respiratory domains~\cite{feldman2022als,schneck2025time}.
This heterogeneity is further reflected in longitudinal measures such as
the ALS Functional Rating Scale--Revised (ALSFRS-R), where patients
demonstrate markedly different trajectories even at similar baseline
stages~\cite{cedarbaum1999alsfrsr}.
Although ALS is universally progressive, the timing, sequence, and
manifestation of clinically meaningful milestones---such as loss of
ambulation, dysphagia requiring gastrostomy, or respiratory
insufficiency necessitating ventilatory support---vary widely across
individuals~\cite{arguedas2025risk,prentice2026semicompeting,chio2015staging}.
These intermediate events represent critical transitions in disease
burden and healthcare utilization, often preceding terminal outcomes and
directly impacting quality of life, caregiver needs, and clinical
management strategies~\cite{prentice2026semicompeting,miller2009practice}.
However, predicting when such events will occur remains challenging due
to the complex and nonlinear nature of disease progression~\cite{bourke2006ventilation,dalgic2021mapping,ward2025wheelchair}.

Existing modeling efforts have primarily focused on survival or composite
endpoints, with limited ability to capture the dynamic and individualized
nature of disease progression.
While approaches such as landmark modeling and semi-competing risks have
improved time-to-event prediction by incorporating longitudinal
information, they remain largely population-level tools and are not
explicitly designed for real-time, patient-specific
prediction~\cite{arguedas2025semicompeting,putter2007tutorial}.
In parallel, machine learning--based prognostic models have demonstrated
improved predictive performance by leveraging high-dimensional clinical
and molecular data; however, many of these models operate as static
predictors and often lack interpretability and temporal adaptability
required for clinical deployment~\cite{vanderburgh2017deep,vieira2022machinelearning,pancotti2022deeplearning}.
These approaches vary substantially in their treatment of time and interpretability: some incorporate longitudinal observations and nonlinear relationships, whereas others rely primarily on baseline variables, summary measures, or fixed prediction horizons. Consequently, dynamic updating of patient-specific risk and explicit modeling of interactions among functional domains and intermediate clinical events remain inconsistent across studies. Joint models, longitudinal subgroup analyses, and deep learning approaches have begun to address these limitations, although challenges remain in external generalizability, clinical interpretability, and multimodal integration~\cite{faghri2022clinicalsubgroups,rizopoulos2012jointmodels,ioannidis2005why,collins2015tripod,steyerberg2009clinical}.
Consequently, there is a critical need for next-generation modeling
frameworks that can capture individual-level disease trajectories,
integrate heterogeneous data streams, and provide reliable,
time-resolved predictions of clinically relevant events to support both
precision medicine and adaptive clinical trial design.

The concept of a digital twin---a dynamic, computational representation
of an individual that evolves with incoming data---provides a promising
framework to address current limitations in modeling ALS
progression~\cite{tao2018digitaltwin,bjornsson2019digitaltwins,corralacero2020digitaltwin,metayer2026data}.
Originating from engineering and cyber-physical systems, digital twins
have been increasingly applied in healthcare to enable real-time
monitoring, prediction, and simulation of patient-specific
trajectories~\cite{bruynseels2018ethical}.
In ALS, integrating longitudinal functional assessments, demographic characteristics, and clinical events may support individualized prediction of disease progression and healthcare utilization.
By incorporating heterogeneous data sources and capturing temporal
dynamics, a good predictive model could extend beyond static prediction by
enabling continuous updating of risk profiles and simulation of future
disease trajectories under varying clinical scenarios~\cite{sun2023healthcare}.
Advances in machine learning, multimodal data integration, and mechanistic modeling further support the development of scalable and interpretable systems~\cite{topol2019highperformance}.
However, despite growing interest and successful applications in cardiovascular disease and oncology, temporal machine-learning approaches to time-to-event prediction remain underexplored in neurodegenerative diseases, particularly for linking longitudinal functional decline with clinical outcomes in ALS~\cite{schork2015personalized}.
This gap underscores the need for time-aware, patient-specific predictive frameworks that generate clinically actionable forecasts to support precision care and proactive intervention.

In this study, we develop a temporal machine learning-based time-to-event (TTE) model for ALS
that integrates longitudinal ALSFRS-R trajectories with survival modeling
to predict clinically relevant healthcare utilization events.
We first construct a harmonized dataset by cross-linking ALS diagnosis
records, longitudinal functional assessments, activities of daily living
(ADL), and demographic data, followed by rigorous filtering to ensure
temporal consistency and data completeness, yielding a high-quality
analytic cohort.
We then identify data-driven functional domains using correlation-based
clustering of ALSFRS-R items, revealing coherent structures across
bulbar, limb, axial, and respiratory functions.
Longitudinal progression is characterized using generalized additive
mixed models (GAMMs), capturing nonlinear and domain-specific patterns
of functional decline.
Additionally, we develop a temporal machine learning framework to model stage-dependent functional transitions across longitudinal visits, enabling individualized prediction of future disease trajectories and capturing the evolving dynamics of ALS progression over time.
Building on these dynamics, we apply a Cox proportional hazards model to
quantify the association between functional impairment and time to healthcare
utilization, demonstrating that lower limb decline---particularly walking
and stair-climbing ability---is the strongest predictor of earlier
assistive device use.
Finally, we implement a this prototype that integrates
patient-specific features to generate individualized survival curves and
enable dynamic prediction of wheelchair-free survival.
Together, this unified framework provides a scalable, interpretable, and
clinically actionable approach for linking disease progression with
personalized decision support in ALS.

\section{Methods}
\label{sec:methods}

\subsection{Data Preprocessing}
\label{sec:data}

We generated a cross-linked dataset using the ALS Natural History
Consortium (ALSNHC), integrating four data files: diagnosis records, ALS
diagnostic history, longitudinal ALSFRS-R assessments, and assistive
device logs.
Unique patient identifiers (SubjectUID) were used to harmonize records
across datasets.

Diagnosis records were loaded to obtain subject identifiers, diagnosis
dates (\texttt{cdalsdt}), and ALS phenotype labels (\texttt{cdalsphn}).
Phenotype codes were retained as six predefined categories: classical ALS
with balanced upper and lower motor neuron involvement, upper motor
neuron--predominant disease, lower motor neuron--predominant disease,
progressive bulbar palsy, primary lateral sclerosis, and progressive
muscular atrophy.
ALS diagnostic history records were processed to derive regional
topographical involvement.
For each anatomical region (bulbar, left and right upper extremity,
trunk, left and right lower extremity), the corresponding upper motor
neuron, lower motor neuron, and electrophysiological evidence fields were
collapsed into a binary region-level indicator.
A region was coded as affected if any of its component indicators equaled
1.
To establish temporality relative to diagnosis, subjects with missing
diagnosis date or follow-up visit date were excluded.
For the remaining records, time since diagnosis at the visit level
($t_{\text{visit\_since\_dx}}$) was calculated as the interval between
visit date and diagnosis date, expressed in months (days divided by
30.4).

Longitudinal ALSFRS-R data were loaded and transformed into domain-level
severity measures.
Domain-specific percentages were generated for bulbar, upper extremity,
axial, lower extremity, respiratory, and overall function.
Records with missing overall percentage were removed.
For item 5, subitems \texttt{ALSFRS5a} and \texttt{ALSFRS5b} were
harmonized into a single derived variable (\texttt{ALSFRS5}) according
to the branching logic defined by the original item 5 response.
ALSFRS-R records were then merged with the filtered diagnosis dataset by
SubjectUID.
For each follow-up record, functional follow-up time since diagnosis
($t_{\text{FU\_since\_dx}}$) was computed from the interval between
ALSFRS follow-up date and diagnosis date in months.
Assistive device log data were loaded to identify clinically meaningful
mobility events.
Records with non-missing device dates were retained; wheelchair-related
events were defined as device codes 4 or 6.
Event time since diagnosis ($t_{\text{dev\_since\_dx}}$) was calculated
in months.
Finally, demographic data were merged by SubjectUID, and age at each
functional assessment was calculated from date of birth and follow-up
date, converted to years by dividing days by 365.25.

\subsection{Correlation-Based Feature Clustering and Aggregated ALSFRS-R Derivation}
\label{sec:clustering}

To quantify relationships among ALSFRS-R items and derive data-driven
aggregated ALSFRS-R scores, we performed a correlation-based clustering analysis
using Pearson correlation coefficients (PCC).
Each ALSFRS-R item is scored on a 5-point ordinal scale (0--4), where 0
indicates the most severe functional impairment and 4 represents normal
function.
Pairwise Pearson correlations and corresponding $p$-values were computed
to construct a correlation matrix.
To reduce noise, correlations below an absolute threshold were set to
zero.

Hierarchical clustering was applied to the correlation matrix using
Ward's linkage method, which minimizes within-cluster variance and
promotes compact, interpretable clusters.
Clusters were determined either by specifying a fixed number of clusters
$k$ or by applying a distance-based cutoff on the dendrogram.
The resulting clustered correlation structure was visualized using
heatmaps with dendrograms, enabling identification of coherent groups of
highly correlated functions.

\subsection{Generalized Additive Mixed Models for Longitudinal ALSFRS-R Dynamics}
\label{sec:gamm}

To evaluate whether changes in ALSFRS-R scores from diagnosis to
follow-up visits are associated with demographic and temporal factors, we
applied a generalized additive mixed modeling (GAMM) framework, which combines flexible spline-based modeling of nonlinear effects with random effects for repeated-measures data \cite{gomezrubio2018generalized,wood2011fast}.
This approach enables flexible modeling of nonlinear temporal effects
while accounting for repeated measurements within individuals.

For each subject $i$ at visit $j$, the longitudinal outcome $Y_{i,j}$
(either the total ALSFRS-R score or individual item/domain scores) was
modeled as:
\begin{equation}
  Y_{i,j} = \beta_0 + \beta_1 \times \mathrm{Age}_{i,j}
    + \beta_2 \times \mathrm{Sex}_i
    + \sum_{k=1}^{df} B_k\!\left(t_{\mathrm{FU\_since\_dx}_{i,j}}\right)
    + \mu_i + \varepsilon_{i,j},
  \label{eq:gamm}
\end{equation}
where $Y_{i,j}$ denotes the observed ALSFRS-R outcome for subject $i$ at
visit $j$; $\beta_0$ is the population-level intercept; $\beta_1$ and
$\beta_2$ denote fixed effects for age at visit and biological sex,
respectively; $B_k(\cdot)$ represents the $k$-th basis function of a
natural cubic spline capturing nonlinear temporal effects; $df$ is the
degrees of freedom controlling spline flexibility;
$\mu_i \sim \mathcal{N}(0, \sigma_\mu^2)$ is a patient-specific random
intercept accounting for within-subject correlation; and
$\varepsilon_{i,j} \sim \mathcal{N}(0, \sigma_\varepsilon^2)$ is the
residual error term.

\subsection{Function Decline Prediction Using a Temporal Machine Learning Model}
\label{sec:ml}
We formulate ALS functional-domain decline as a \emph{multi-domain
discrete-time survival problem} and introduce a novel five-head hazard
model that simultaneously predicts time-to-next-transition for all five
ALSFRS-R domains, building on discrete-time survival modeling, neural survival analysis, and attention-based sequence modeling for irregular longitudinal data\cite{allison1982discrete, kvamme2019time, lee2018deephit, vaswani2017attention}.

\subsubsection{Problem Formulation}

Let a patient be characterised by a sequence of clinical visits
$\mathcal{H}_i = \{(\mathbf{x}_t, \mathbf{s}_t, \tau_t)\}_{t=1}^{T_i}$,
where $\mathbf{x}_t \in \mathbb{R}^{d_s}$ denotes static demographic
features (identical across visits),
$\mathbf{s}_t \in \mathbb{R}^{d_v}$ denotes longitudinal visit
measurements, and $\tau_t$ is the calendar date of visit $t$.
For each patient we define five domain scores $y_t^{(d)}$:

\begin{equation}
y_t^{(d)} \in \mathbb{R}_{\geq 0},
d \in \{\text{bulbar, fine\_motor, gross\_motor, walking, respiratory}\},
\end{equation}

computed from subsets of ALSFRS-R items.

\textbf{Event definition.}
Intuitively, a domain transition event marks the first clinically detectable worsening within a functional domain after the landmark (index) visit operationally, a drop of at least one point on that domain's ALSFRS-R subscale.For domain $d$ at landmark visit $t_0$, the \emph{domain transition
event} $E_d$ is the first future visit $t > t_0$ at which
$y_t^{(d)} \leq y_{t_0}^{(d)} - 1$.
If no such visit is observed (patient lost to follow-up or drop not
reached), the observation is \emph{right-censored} at the last available
follow-up.

\textbf{Discrete-time formulation.}
We discretise the time axis into $K = 24$ intervals of width
$\Delta = 30$ days, yielding a prediction horizon of 720 days
(approximately 2 years).
For event $E_d$, let $T_d$ denote the event time in days and let
$b_d = \lfloor T_d / \Delta \rfloor$ be the corresponding bin index,
clamped to $[0, K{-}1]$.
The per-interval hazard is
$h_{d,k} = P(b_d = k \mid b_d \geq k, \mathcal{H})$, and the survival
function is:
\begin{equation}
  S_d(k) = \prod_{j=0}^{k-1}(1 - h_{d,j}).
  \label{eq:survival}
\end{equation}
The goal is to learn a function
$f_\theta(\mathcal{H}_i) = \{h_{d,k}\}_{d,k}$ that maps patient history
to per-domain hazard curves.

\subsubsection{Model Architecture}
\label{sec:model}

The Attention-based Stage Transition Predictor (ASTP) model comprises four components.

\paragraph{Static Embedding}
Static features $\mathbf{x} \in \mathbb{R}^{d_s}$ are embedded via a
two-layer MLP followed by LayerNorm and ReLU:
\begin{equation}
  \mathbf{e}_s = \text{ReLU}(\text{LN}(W_s \mathbf{x} + b_s)) \in \mathbb{R}^{d}.
  \label{eq:static_embed}
\end{equation}
This embeds the fixed patient demographics into the model dimension
$d = 128$.

\paragraph{Transformer Sequence Encoder}
Visit sequences are projected to model dimension via a linear layer and
LayerNorm, then processed by a standard Transformer encoder with
pre-norm (``Pre-LN'') residual connections~\cite{xiong2020layer}:
\begin{equation}
  \mathbf{Z} = \text{TransformerEncoder}(\text{LN}(W_v \mathbf{S})) \in \mathbb{R}^{T \times d}.
  \label{eq:encoder}
\end{equation}
The encoder consists of $L = 3$ layers, each with $H = 4$ attention
heads, a feedforward dimension of $4d$, GELU activations, and dropout
$p = 0.1$.
Sequences are left-padded to the batch maximum length; a per-patient
Boolean padding mask ($\texttt{True} = \text{ignore}$) is passed to all
attention operations in PyTorch's convention. Temporal ordering and irregular spacing are encoded explicitly through the elapsed-time input features (months since diagnosis and days since the previous visit) rather than through sinusoidal positional encodings, which assume regular sampling; cross-visit dependencies are then captured by the self-attention layers.

\paragraph{Cross-Attention Pooling}
To aggregate the variable-length encoded sequence into a fixed-size
patient state, we use cross-attention with a single learnable query
vector:
\begin{equation}
  \mathbf{c} = \text{MultiheadAttn}(\mathbf{q}, \mathbf{Z}, \mathbf{Z}) \in \mathbb{R}^{d},
  \label{eq:pool}
\end{equation}
where $\mathbf{q} \in \mathbb{R}^{d}$ is learned jointly with the model.
The static embedding and context vector are then fused:
\begin{equation}
  \mathbf{u} = \text{ReLU}(\text{LN}(W_f [\mathbf{e}_s \| \mathbf{c}])) \in \mathbb{R}^{d},
  \label{eq:fusion}
\end{equation}
where $\|$ denotes concatenation.

\paragraph{Domain-Specific Hazard Heads}
For each domain $d$, a two-layer MLP receives the fused representation
augmented by the current normalised domain score
$\tilde{y}_d = y_{t_0}^{(d)} / y_{\max}^{(d)} \in [0,1]$:
\begin{equation}
  \mathbf{h}_d = \sigma\!\left(W_d^{(2)} \text{ReLU}(W_d^{(1)} [\mathbf{u} \| \tilde{y}_d] + b_d^{(1)}) + b_d^{(2)}\right) \in [0,1]^K,
  \label{eq:head}
\end{equation}
where $\sigma$ is the element-wise sigmoid function and $K = 24$ is the
number of time bins.
Conditioning on the current score is critical: a domain near its
maximum (no room to drop) behaves differently from one near the floor,
where further decline is impossible.
The domain heads are fully independent---they do not share parameters
beyond the fused representation $\mathbf{u}$.

The total model has approximately 749K trainable parameters.

\subsubsection{Inference and 5-Year Extrapolation}
\label{sec:infer}

At inference time, for each domain $d$ the survival curve is computed via~\eqref{eq:survival} over the model horizon of 720 days.
To provide a clinically useful 5-year view, we extrapolate beyond 720 days by holding the hazard constant at the mean of the last three predicted hazard bins:
\begin{equation}
  h_{d,k}^\text{ext} = \frac{1}{3}\sum_{j=K-3}^{K-1} h_{d,j}, \quad k \geq K.
\end{equation}
The predicted time to domain transition under probability threshold $\kappa$ is the first day $t^*$ such that $S_d(t^*) < \kappa$:
\begin{equation}
  t_d^*(\kappa) = \inf \{t \geq 0 : S_d(t) < \kappa\},
\end{equation}
which defaults to $\kappa = 0.70$ (i.e., a 30\% cumulative probability of transition has been crossed).
Users of the clinical interface can adjust $\kappa$ interactively.

\subsection{Time-to-Event Modeling Using Cox Proportional Hazards}
\label{sec:cox}

To evaluate the association between longitudinal functional status and
the timing of assistive device utilization, we applied a Cox
proportional hazards regression model, a standard semiparametric survival modeling framework for estimating covariate effects on TTE outcomes \cite{cox1972regression}.
The primary outcome was defined as the time to device use (e.g.,
wheelchair), $t_{\text{dev\_since\_dx}}$, representing the time in
months from diagnosis to the recorded device usage event. Gastrostomy and ventilatory support were not modeled because these events, particularly ventilation initiation, were too sparsely and incompletely recorded for robust TTE analysis.

The model incorporated both demographic and longitudinal clinical
features.
Predictor variables included age at visit, follow-up time since
diagnosis ($t_{\text{FU\_since\_dx}}$), sex, and individual ALSFRS-R
item scores (ALSFRS1--ALSFRS9 and ALSFRS1r--ALSFRS3r) capturing
multi-domain functional status.
To account for repeated observations per subject, clustering was
specified at the subject level using SubjectUID, and robust variance
estimation was applied to adjust standard errors for within-subject
correlation.
Visit-level weighting (VisitCount) was incorporated to reflect the
relative contribution of each observation.
Model regularization was implemented using elastic net penalization
($\lambda = 0.05$, $\alpha_{\text{L1}} = 0.2$) to mitigate
multicollinearity among correlated ALSFRS-R predictors~\cite{zou2005regularization}.

The Cox proportional hazards model is:
\begin{equation}
  h_i(t|\mathbf{X}_i) = h_0(t)\exp\!\left(
    \beta_1 t_{\mathrm{Age}_{t|i}}
    + \beta_2 t_{\mathrm{FU}_{t|i}}
    + \beta_3 \mathrm{Sex}_i
    + \sum_{k=1}^{12}\beta_{3+k} \, \mathrm{ALSFRS}_{k,t|i}
  \right),
  \label{eq:cox}
\end{equation}
where $h_0(t)$ is the unspecified baseline hazard, $\mathbf{X}_i$ is the
covariate vector for subject $i$, and $\mathrm{ALSFRS}_{k,t|i}$ denotes the
$k$-th ALSFRS-R item score for subject $i$ at time $t$.

The penalized Cox model estimates $\hat{\beta}$ by maximizing the
penalized partial log-likelihood:
\begin{equation}
  \hat{\beta} = \arg\max_{\beta}\!\left(
    \ell(\beta) - \lambda\!\left[
      \alpha \sum_{j=1}^{p}|\beta_j|
      + \frac{1-\alpha}{2}\sum_{j=1}^{p}\beta_j^2
    \right]
  \right),
  \label{eq:penalized}
\end{equation}
where $\ell(\beta)$ is the Cox partial log-likelihood, $\lambda$ is the
overall penalty strength, and $\alpha$ controls the elastic net mixing
between L1 and L2 penalties ($\lambda = 0.05$, $\alpha = 0.2$).

The partial log-likelihood is:
\begin{equation}
  \ell(\beta) = \sum_{i:\delta_i=1}\!\left[
    x_i^{\top}\beta
    - \log\!\left(\sum_{l \in R(t_i)}\exp(x_l^{\top}\beta)\right)
  \right],
  \label{eq:partial_ll}
\end{equation}
where $\delta_i$ is the event indicator for patient $i$, $t_i$ is the
observed event time, and $R(t_i)$ is the risk set at time $t_i$.

\subsection{Interactive Clinical Decision-Support Application}
\begin{figure}[htbp]
\centerline{\includegraphics[width=\columnwidth]{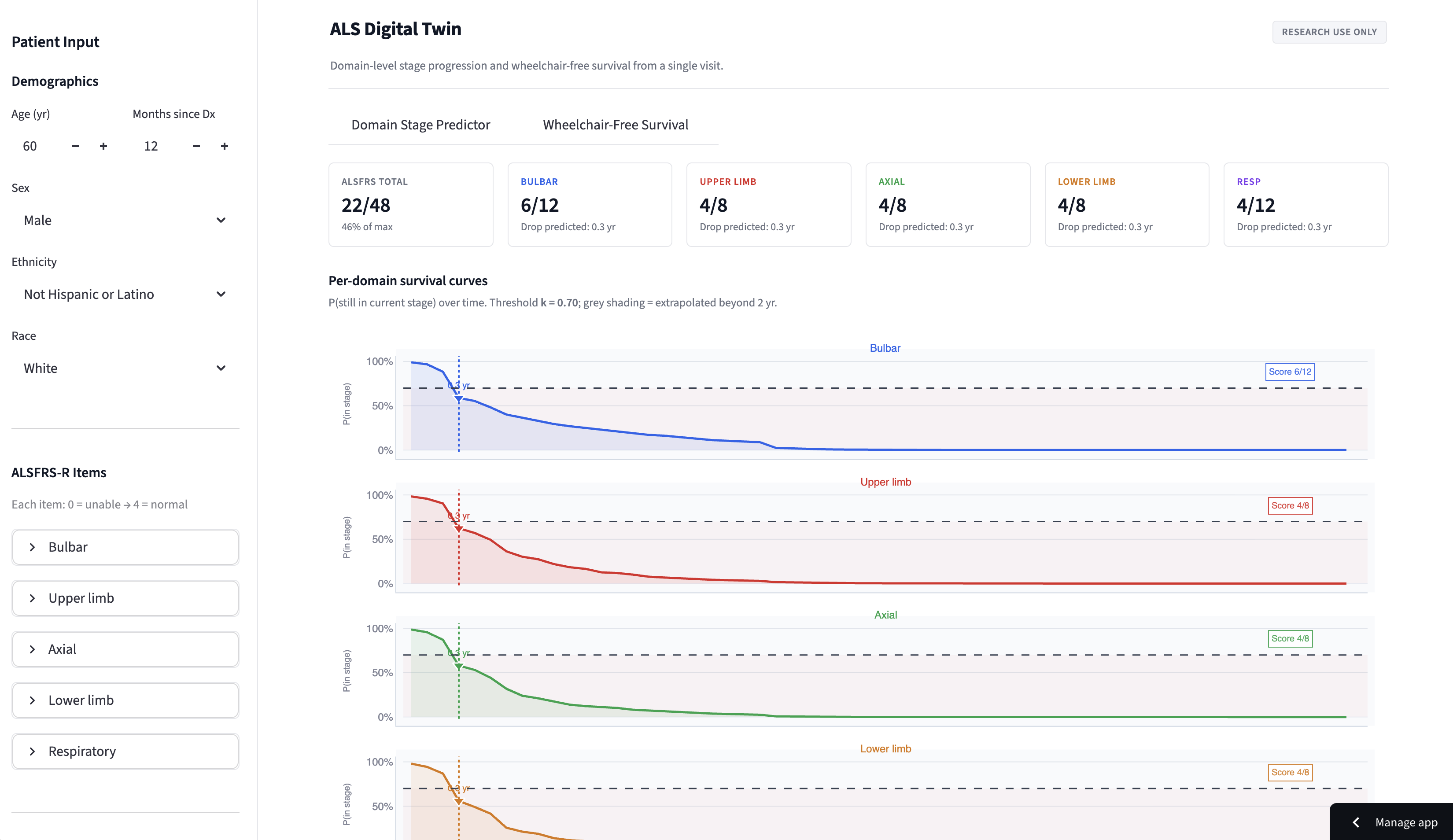}}
\caption{Digital twin-inspired temporal machine learning-based TTE model for domain stage prediction and time-to-event prediction of healthcare
utilization. A Cox-based probabilistic framework generates a personalized
survival curve reflecting the evolving risk of wheelchair-free survival.
In the example shown, a patient with relatively preserved baseline
function has a predicted median time to event of approximately 75.5
months.}
\label{fig:digitaltwin}
\end{figure}

Beyond the underlying models, we deploy the predictive model as a publicly available, open-source Streamlit web application to translate TTE models into interactive clinical decision-support systems while maintaining interpretability and usability \cite{london2019artificial}(Fig.~\ref{fig:digitaltwin}). It translates a single clinic visit into a multi-horizon prognostic readout, lowering the data-entry burden that has
  historically gated bedside adoption of ALS progression models. The interface collects only the routinely-captured ALSFRS-R items.
 Using these inputs, the application drives two complementary back-ends in parallel. The first mode invokes the Transformer-based Domain Stage
  Predictor and renders, per ALSFRS-R domain (bulbar, upper limb, axial, lower limb, respiratory), the survival probability of remaining at
  the current functional stage together with the underlying 30-day-bin hazard rates, the predicted transition time at a user-tunable survival
  threshold $k$, and a four-level risk band ($<$6 mo, $<$1 yr, $<$2 yr, $>$2 yr); to extend the trained 24-bin (720-day) horizon to a five-year
  clinically meaningful window, the application applies a constant-hazard tail extrapolation derived from the final three trained bins, with
  the extrapolated region visually shaded so users can distinguish it from in-distribution predictions. The second mode wraps a Cox
  proportional-hazards model fit to time-to-wheelchair-access ($n = 15$ predictors: age at visit, follow-up since diagnosis, sex, and the 12
  ALSFRS-R items) and reports the wheelchair-free survival curve, the median time-to-event, and discrete five-horizon probabilities (6, 12,
  24, 36, 60 months). All figures are rendered with Plotly for interactive inspection (per-point hover, hazard overlays, threshold sliders),
  and each prediction view is paired with an expandable Model \& Data Details panel that surfaces the architecture, training cohort size,
  feature set, and the research use only qualifier required for non-regulatory deployment. The application caches model weights and scalers
  via Streamlit's resource-caching layer, so inference for a single visit runs in well under a second on commodity CPU hardware, making the
  tool deployable on a clinic laptop without GPU access.

\section{Results}
\label{sec:results}

\subsection{Cohort Construction and Data Integration}
\label{sec:cohort}

\begin{figure}[htbp]
\centerline{\includegraphics[width=\columnwidth]{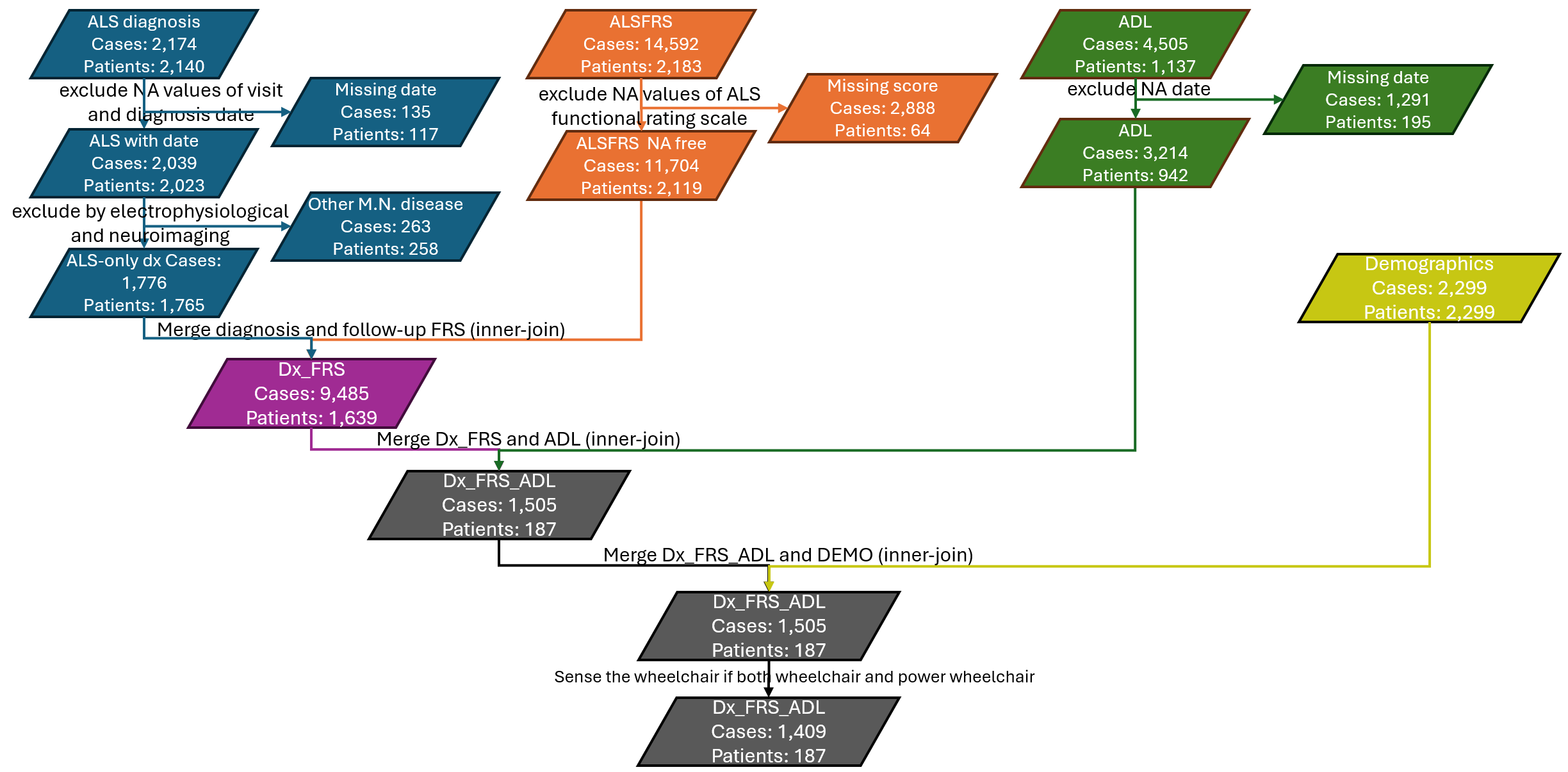}}
\caption{Workflow for cross-linked dataset construction and cohort
selection. Overview of data preprocessing and integration across ALS
diagnosis, ALSFRS-R, ADL, and demographic datasets. Records were
filtered for completeness and merged at the subject level. The final
analytic dataset was derived after sequential filtering and integration
steps.}
\label{fig:workflow}
\end{figure}

Cohort assembly is summarized in Fig.~\ref{fig:workflow}.
After excluding records with missing diagnosis dates and non-ALS motor
neuron diseases, 1,776 ALS-specific cases (1,765 patients) were retained
from the initial diagnosis dataset.
ALSFRS-R data were filtered to remove missing scores, yielding 11,704
observations (2,119 patients), which were subsequently linked to
diagnosis records to form a combined dataset of 9,485 cases (1,639
patients).
ADL records were similarly filtered for completeness and merged,
resulting in 1,505 cases from 187 patients.
Integration with demographic data and additional filtering for
longitudinal completeness produced the final analytic cohort of 1,409
cases from 187 patients.

\subsection{Functional Domain Identification Based on Question-Level Similarity}
\label{sec:domains}

\begin{figure}[htbp]
\centerline{\includegraphics[width=\columnwidth]{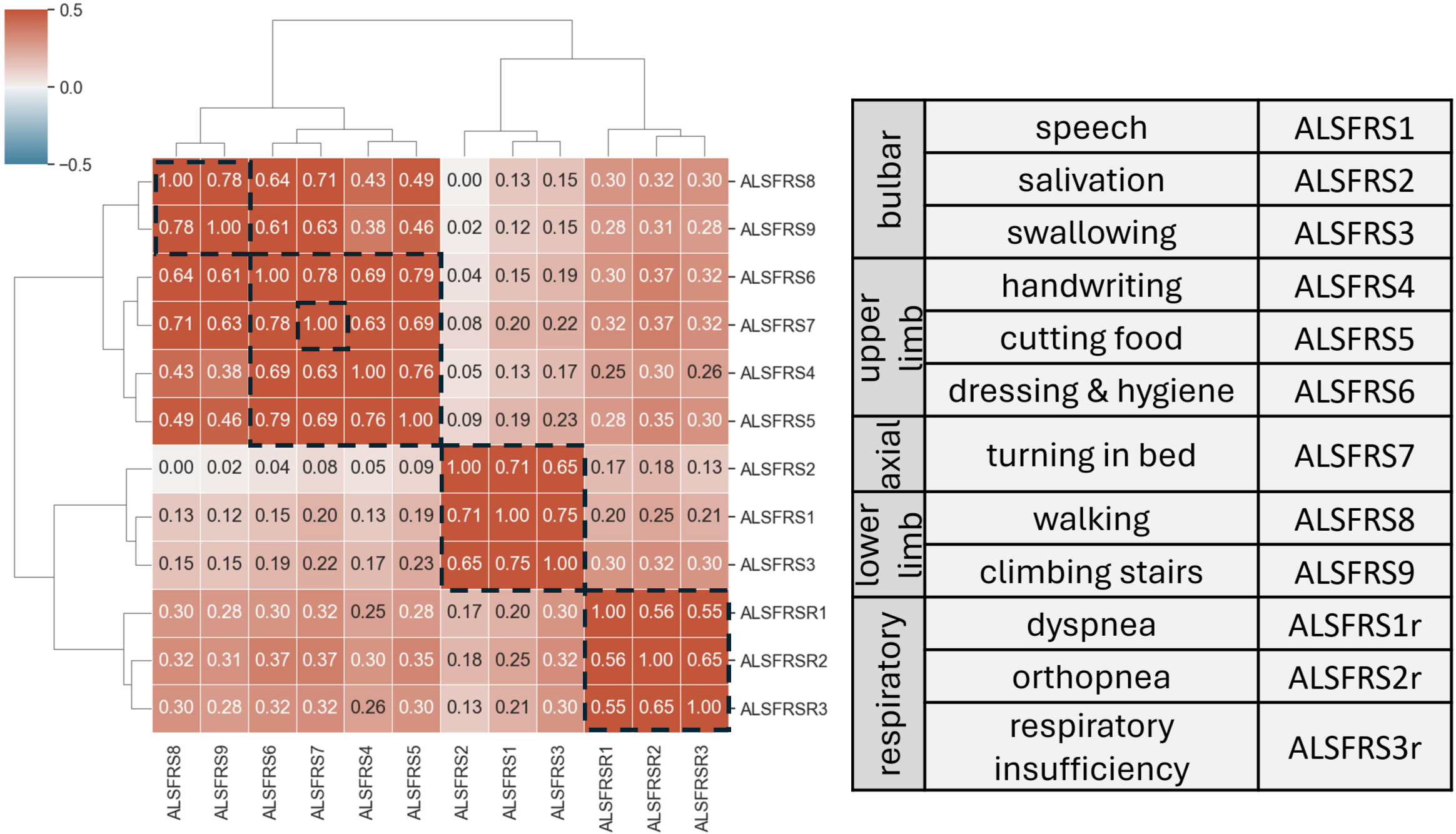}}
\caption{Correlation structure and domain organization of ALSFRS-R items.
Left: Hierarchical clustering heatmap of Pearson correlations among
ALSFRS-R items. Right: Clinical grouping of ALSFRS-R items into
functional domains (bulbar, upper limb, axial, lower limb, respiratory).}
\label{fig:correlation}
\end{figure}

To examine the internal structure of ALSFRS-R items and validate
domain-level groupings, we performed Pearson correlation analysis
followed by hierarchical clustering using Ward's linkage
(Fig.~\ref{fig:correlation}).
The clustered heatmap revealed strong within-domain correlations and
clear separation across functional domains.
Items related to limb function (ALSFRS4--ALSFRS9) formed a tightly
correlated cluster, indicating coordinated decline in motor function.
A minimum pairwise Pearson correlation threshold of $\geq 0.7$ supported
subdivision into upper limb (ALSFRS4--ALSFRS6), axial (ALSFRS7), and
lower limb (ALSFRS8--ALSFRS9) components.
Bulbar items (ALSFRS1--ALSFRS3) exhibited high inter-item correlations
and clustered distinctly, reflecting shared cranial nerve--mediated
impairments.
Similarly, respiratory items (ALSFRSR1--ALSFRSR3) demonstrated strong
internal correlations and formed a separate cluster.
Cross-domain correlations were comparatively weaker, supporting the
relative independence of functional systems.

\subsection{Longitudinal Effects on ALSFRS-R Functional Decline}
\label{sec:longitudinal}

Generalized additive mixed modeling revealed a consistent and significant
decline in ALSFRS-R item scores over time across all functional domains
(Table~\ref{tab:gamm}).
The nonlinear time effects were highly significant ($p < 0.001$ for most
components) and uniformly negative, indicating progressive functional
deterioration from diagnosis.
The magnitude of decline was particularly pronounced in upper and lower
limb functions (ALSFRS4--ALSFRS9) and respiratory measures.

Sex effects were observed in selected domains, with significant
associations in handwriting (ALSFRS4, $p = 0.002$) and dressing (ALSFRS6, $p = 0.016$), as well as in respiratory orthopnea (ALSFRSR2,
$p = 0.041$).
In contrast, age at visit showed limited influence on functional
outcomes, with only salivation (ALSFRS2, $p = 0.024$) exhibiting a
modest age-associated decline.

\begin{table}[htbp]
\centering
\caption{Longitudinal effects of time, sex, and age on ALSFRS-R item-level
outcomes from GAMMs. $\downarrow$ indicates decline over time; (***) $p<0.001$,
(**) $p<0.01$, (*) $p<0.05$, ns = not significant.}
\label{tab:gamm}
\begin{tabular}{llllccc}
\toprule
\textbf{Domain} & \textbf{Function} & \textbf{Var.} & \textbf{Time} & \textbf{Sex} & \textbf{Age} \\
\midrule
\multirow{3}{*}{\textbf{Bulbar}}
  & Speech        & ALSFRS1  & $\downarrow$(***)  & ns              & ns \\
  & Salivation    & ALSFRS2  & $\downarrow$(**)   & ns              & $\downarrow$(*) \\
  & Swallowing    & ALSFRS3  & $\downarrow$(***)  & ns              & ns \\
\multirow{3}{*}{\textbf{Upper limb}}
  & Handwriting   & ALSFRS4  & $\downarrow$(***)  & $\downarrow$(**) & ns \\
  & Cutting food  & ALSFRS5  & $\downarrow$(**)   & ns              & ns \\
  & Dress.\&hyg.  & ALSFRS6  & $\downarrow$(***)  & $\downarrow$(*) & ns \\
\textbf{Axial}
  & Turn. in bed  & ALSFRS7  & $\downarrow$(***)  & ns              & ns \\
\multirow{2}{*}{\textbf{Lower limb}}
  & Lower limb    & ALSFRS8  & $\downarrow$(***)  & ns              & ns \\
  & Cl. stairs    & ALSFRS9  & $\downarrow$(***)  & ns              & ns \\
\multirow{3}{*}{\textbf{Respiratory}}
  & Dyspnea       & ALSFRSR1 & $\downarrow$(**)   & ns              & ns \\
  & Orthopnea     & ALSFRSR2 & $\downarrow$(**)   & $\downarrow$(*) & ns \\
  & Resp.\ insuf. & ALSFRSR3 & $\downarrow$(**)   & ns              & ns \\
\bottomrule
\end{tabular}
\end{table}

\subsection{Stage Predictor: Single-Patient Case Study}
\label{sec:stage_demo}

We evaluate the multi-domain stage predictor in two complementary
ways: (i) qualitative inspection of per-patient trajectory and
survival-curve outputs (\S\ref{sec:stage_demo}--\S\ref{sec:visit_groups}),
and (ii) quantitative summaries of per-domain hazard discrimination
across the held-out test set (\S\ref{sec:ml}--\S\ref{sec:model}).
The two views correspond to different evaluation units (patient-level
trajectory MAE in ALSFRS-R points vs.\ domain-level event-time MAE in
days and concordance via the C-index) and therefore report on
different scales.

\begin{figure}[htbp]
\centerline{\includegraphics[width=\columnwidth]{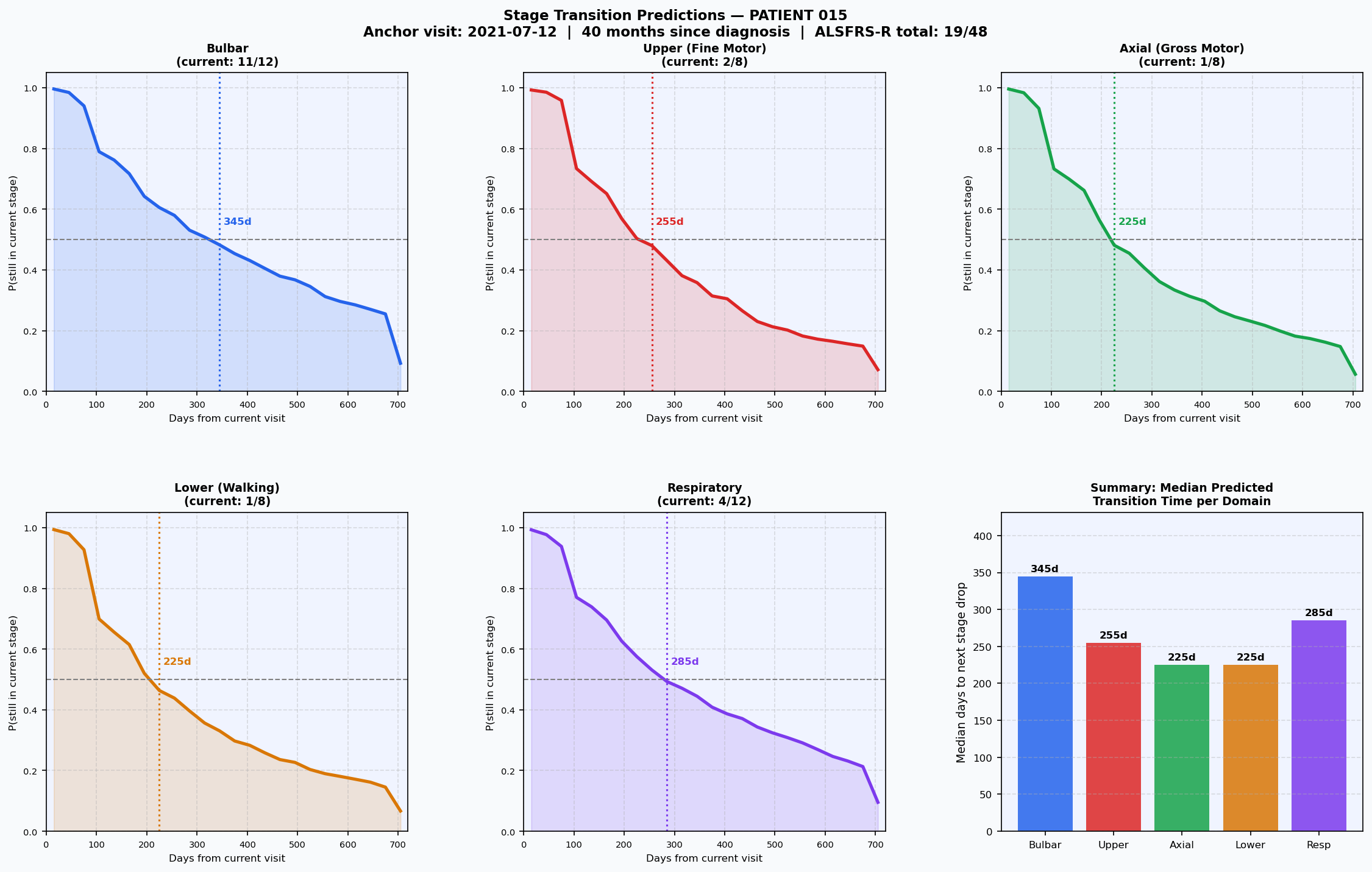}}
\caption{ASTP model output for a representative held-out
test patient (\texttt{PATIENT 015}; female, age 68 at diagnosis,
9 longitudinal visits, ALSFRS-R total 19/48 at the anchor visit
40 months post-diagnosis). Top row: discrete-time survival curves
$S(t) = \Pr(\text{still in current stage})$ for the bulbar, upper
limb, and axial domains across the 720-day model horizon;
bottom-left two panels: lower limb and respiratory curves.
Bottom-right: per-domain median predicted time to a $\geq$1-point
drop. Dashed line marks the 50\% survival threshold; dotted vertical
lines mark the predicted median transition time per domain.}
\label{fig:stage_demo}
\end{figure}

\begin{figure}[htbp] 
\centering 
\subfloat[Rapid progressor.]{
  \includegraphics[width=0.45\textwidth]{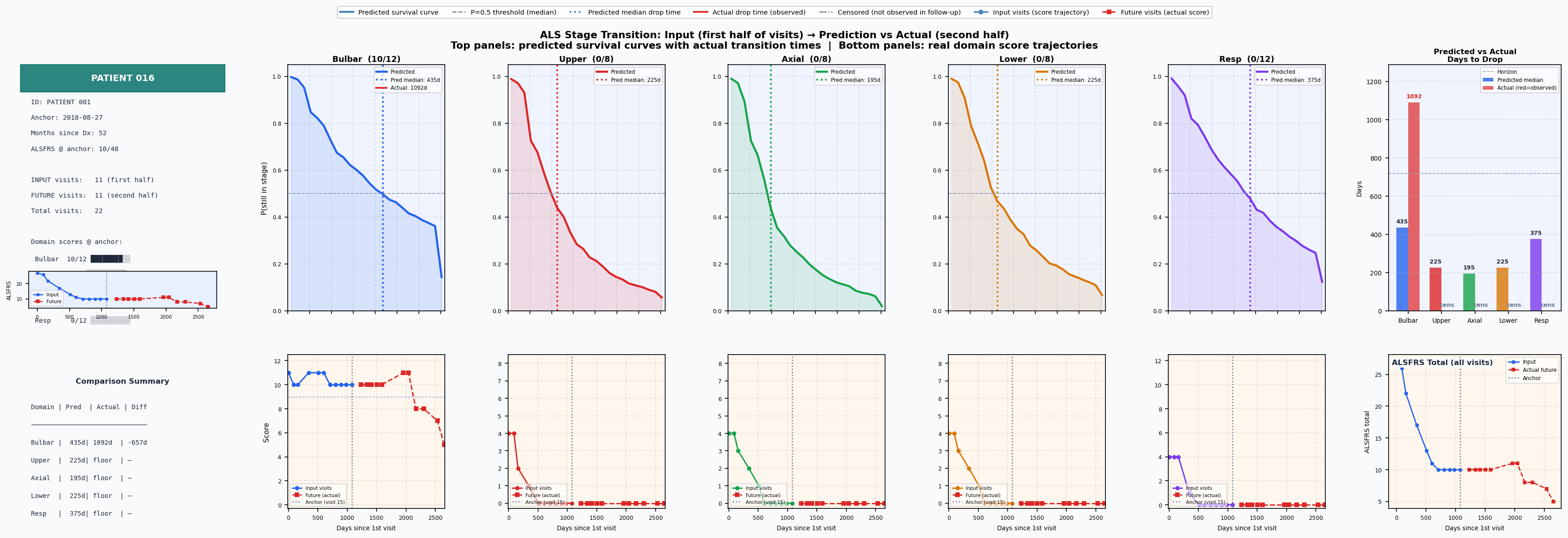}
  \label{fig:survivalcurves_rapid}
}
\hfill
\subfloat[Slow progressor.]{
  \includegraphics[width=0.45\textwidth]{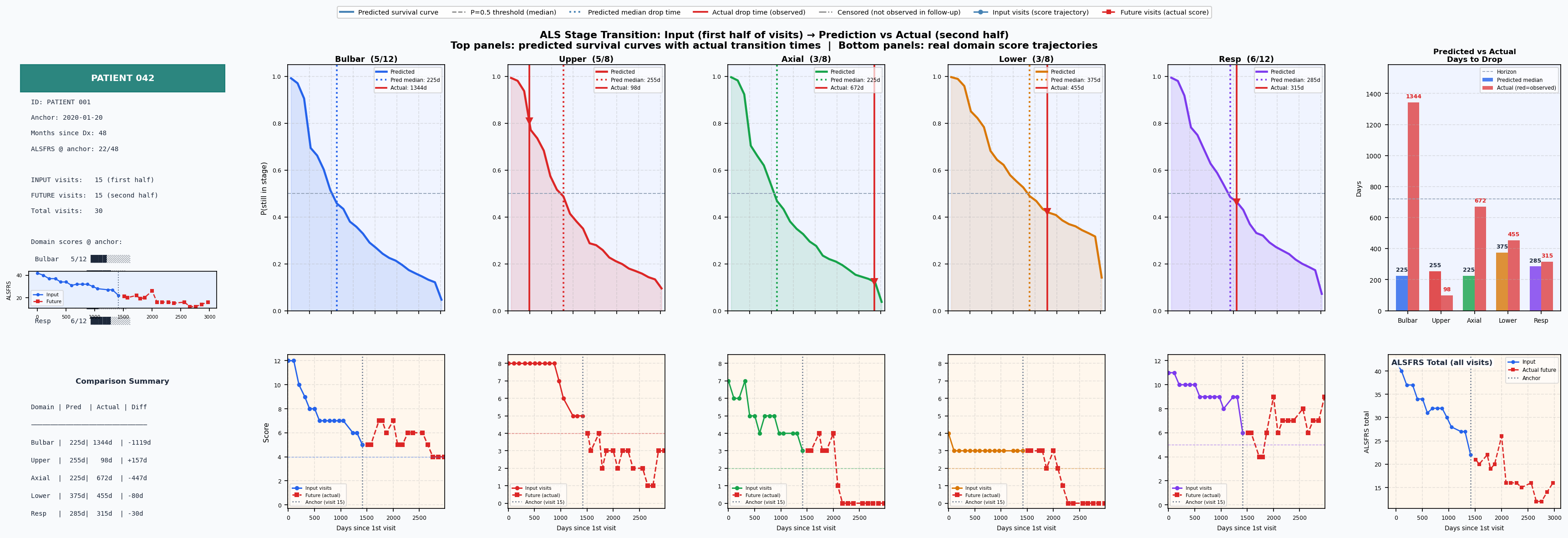}
  \label{fig:survivalcurves_slow}
}
\hfill
\subfloat[Moderate progressor.]{
  \includegraphics[width=0.45\textwidth]{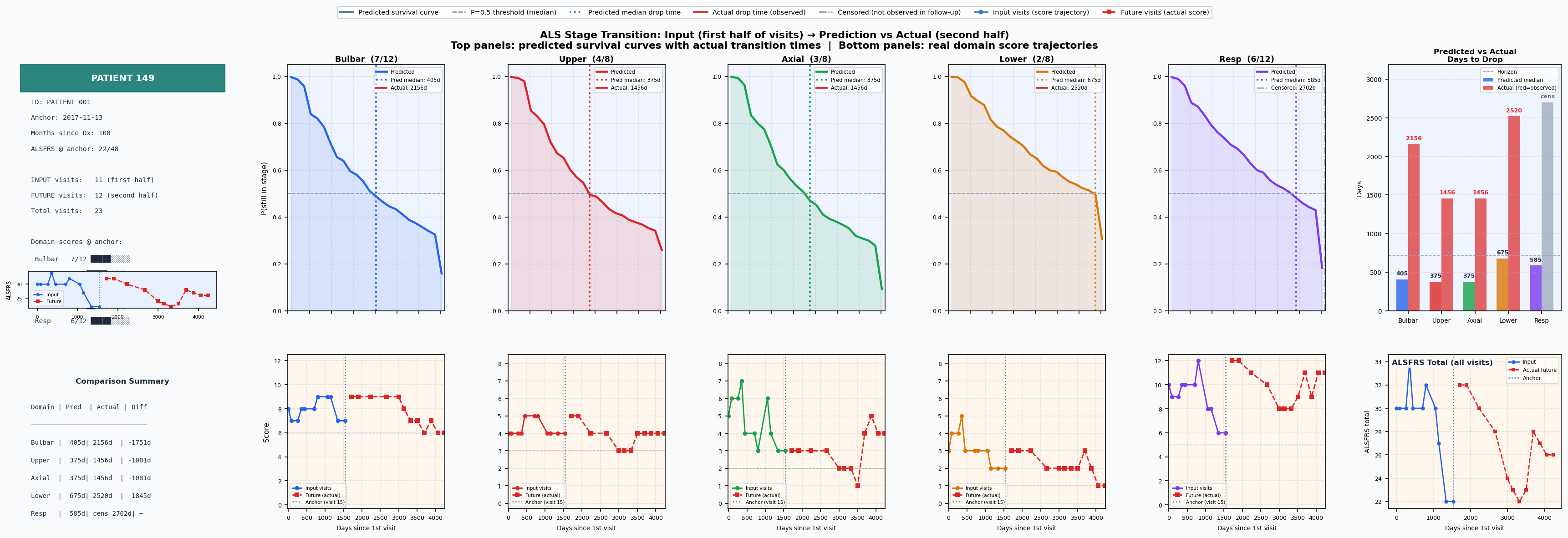}
  \label{fig:survivalcurves_moderate}
}

\caption{Predicted 5-year domain survival curves for three representative ALS patients: rapid, slow, and moderate progressors. Solid lines show predicted survival probabilities, shaded regions indicate quartile prediction intervals, and vertical dashed lines mark the predicted transition time at $\kappa = 0.70$.}
\label{fig:survivalcurves}
\end{figure}

To illustrate the per-patient output of the ASTP model, we
applied the trained model to a held-out test patient
(\texttt{PATIENT 015}) using all visits up to the most recent
landmark as input (Fig.~\ref{fig:stage_demo}).
For each of the five functional domains the model emits a hazard vector
$h_d(t)$ over 24 monthly bins, from which the discrete-time survival
function $S_d(t) = \prod_{u \leq t}(1 - h_d(u))$ is obtained and three
clinically interpretable milestones are read off the curve:
the 25\% transition time (first bin at which $S_d(t)<0.75$),
the median (first bin at which $S_d(t)<0.50$),
and the 75\% transition time (first bin at which $S_d(t)<0.25$).
In this case the heterogeneity the model is designed to capture is
clearly expressed: the bulbar domain---still scoring 11/12 at the
anchor---retains a near-flat survival curve with a median
transition time of 345 days, whereas the three motor domains
(upper limb 2/8, axial 1/8, lower limb 1/8), which are already close
to the floor, enter steeply declining hazard trajectories with
median transition windows of only 225--255 days; the respiratory
domain (4/12) sits between these extremes at 285 days.
Each predicted milestone is paired with the corresponding
ground-truth visit, when available, allowing direct visual
verification of whether the observed score drop fell inside the
predicted credible window.
Fig.~\ref{fig:survivalcurves} complements this single-patient view
with three additional representative held-out trajectories spanning
rapid, moderate, and slow progressors.

\subsection{Cohort-Wide Decline Projection to Floor Score}
\label{sec:decline_to_zero}

\begin{figure*}[t]
\centering

\subfloat[\emph{Short} group --- patient \texttt{PATIENT 024}, 2 visits, MAE = 0.44.]{
  \includegraphics[width=0.48\textwidth]{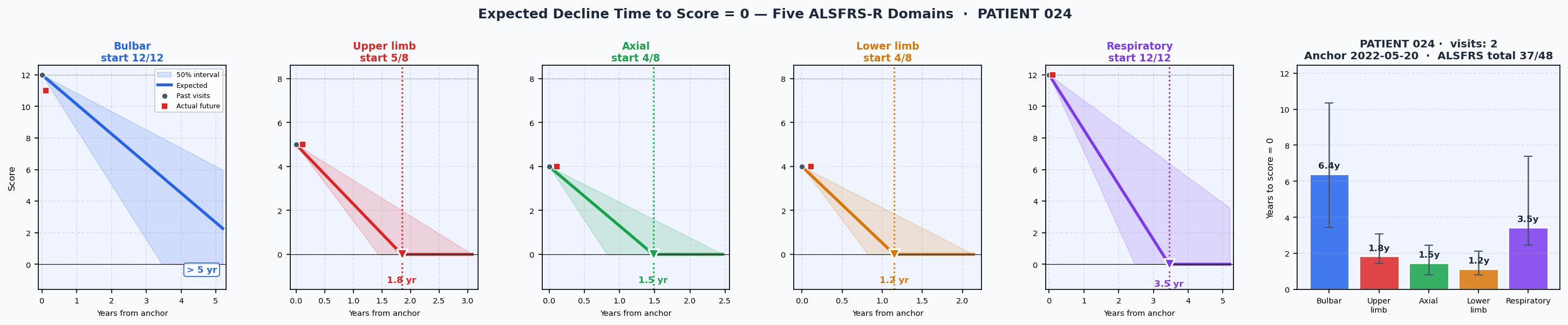}
  \label{fig:decline_short}
}
\hfill
\subfloat[\emph{Medium} group --- patient \texttt{PATIENT 047}, 4 visits, MAE = 0.30.]{
  \includegraphics[width=0.48\textwidth]{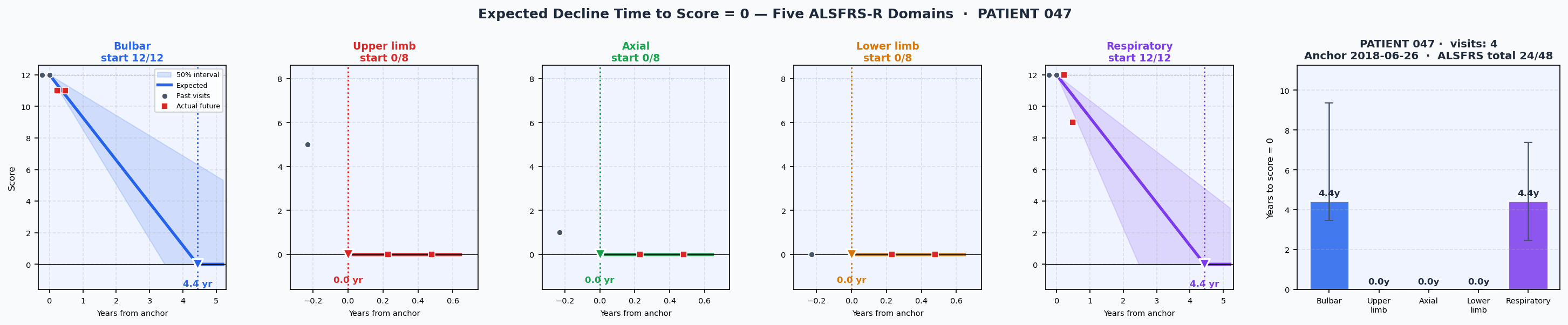}
  \label{fig:decline_medium}
}

\vspace{0.6em}

\subfloat[\emph{Long} group --- patient \texttt{PATIENT 062}, 8 visits, MAE = 0.63.]{
  \includegraphics[width=0.48\textwidth]{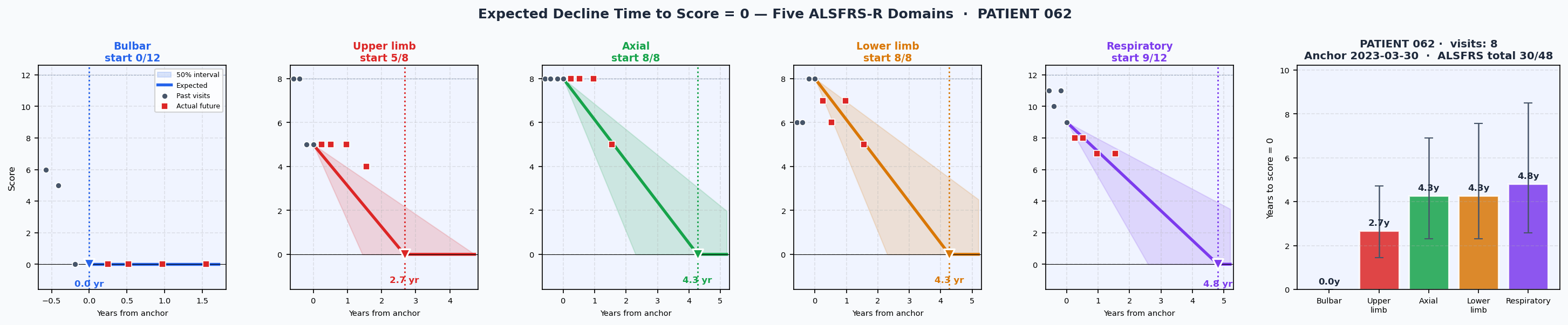}
  \label{fig:decline_long}
}
\hfill
\subfloat[\emph{Extended} group --- patient \texttt{PATIENT 016}, 22 visits, MAE = 0.24.]{
  \includegraphics[width=0.48\textwidth]{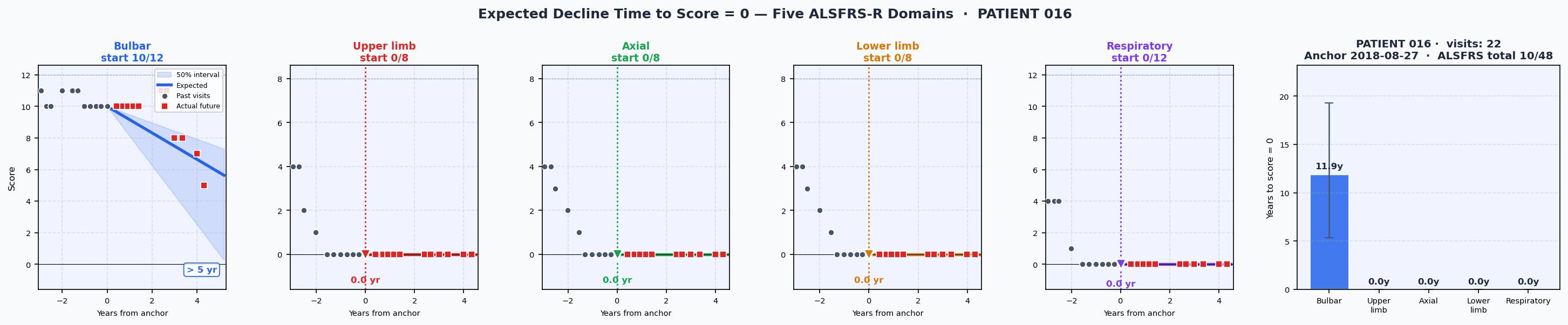}
  \label{fig:decline_extended}
}

\caption{Best-case cohort-wide decline-to-zero projections, one representative patient per visit-count group: Short, Medium, Long, and Extended. For each ALSFRS-R domain, the expected trajectory (solid line) and 50\% prediction interval (shaded band) are projected forward from the anchor visit until the score is expected to reach zero. Grey dots indicate past input visits; red squares indicate held-out future visits. The right-most panel of each subfigure summarizes the per-domain time-to-zero with error bars spanning the 25\%--75\% interval.}
\label{fig:decline_to_zero}
\end{figure*}

We extended the single-patient analysis to the full cohort by running
the ASTP model on every subject with $\geq 2$ recorded
visits across the train, validation, and test splits, yielding 173
evaluable patients.
For each patient, the first half of their visit sequence was used as
model input and the remaining half was withheld as ground truth.
Because the trained model horizon is 720 days, hazard rates were
extrapolated to a 5-year window by carrying the mean of the final
three bins as a constant tail, after which the predicted survival
curve was inverted to obtain three quantiles of time-to-floor-score
($T_{25\%}$, $T_{50\%}$, $T_{75\%}$) per domain by scaling the
predicted slope by the current score.
Fig.~\ref{fig:decline_to_zero} shows the best-fitting case from each
of the four visit-count groups (Short, Medium, Long, Extended; see
\S\ref{sec:visit_groups} for group definitions), demonstrating that
the model's expected decline trajectory and 50\% prediction band
closely track the held-out actual visits across the full range of
follow-up densities.
In the \emph{Short} group (Fig.~\ref{fig:decline_short}), a single
input visit is sufficient to anchor a near-perfect short-horizon
projection (MAE 0.44 across all five domains).
In the \emph{Medium} (Fig.~\ref{fig:decline_medium}) and \emph{Long}
(Fig.~\ref{fig:decline_long}) groups, the held-out trajectories sit
inside the 50\% prediction band for every domain, including domains
that remain well above the floor at the end of follow-up
(annotated as $>5$~yr).
The \emph{Extended} case (Fig.~\ref{fig:decline_extended}, 22 visits
over $\approx$4.3 years) achieves the lowest per-patient error in
the cohort (MAE 0.24) and demonstrates that the constant-hazard
tail extrapolation remains faithful to a multi-year slow-progressor
trajectory.
Across the four sub-figures, the per-patient MAE decreases
monotonically with follow-up density
(0.44 $\rightarrow$ 0.30 $\rightarrow$ 0.63 $\rightarrow$ 0.24),
mirroring the cohort-level monotonic improvement reported in
\S\ref{sec:visit_groups} (the apparent non-monotonicity at the
\emph{Long} representative reflects that we display each group's
\emph{single best} case rather than its mean).

\subsection{Projection Accuracy Stratified by Follow-up Density}
\label{sec:visit_groups}

\begin{figure}[htbp]
\centerline{\includegraphics[width=\columnwidth]{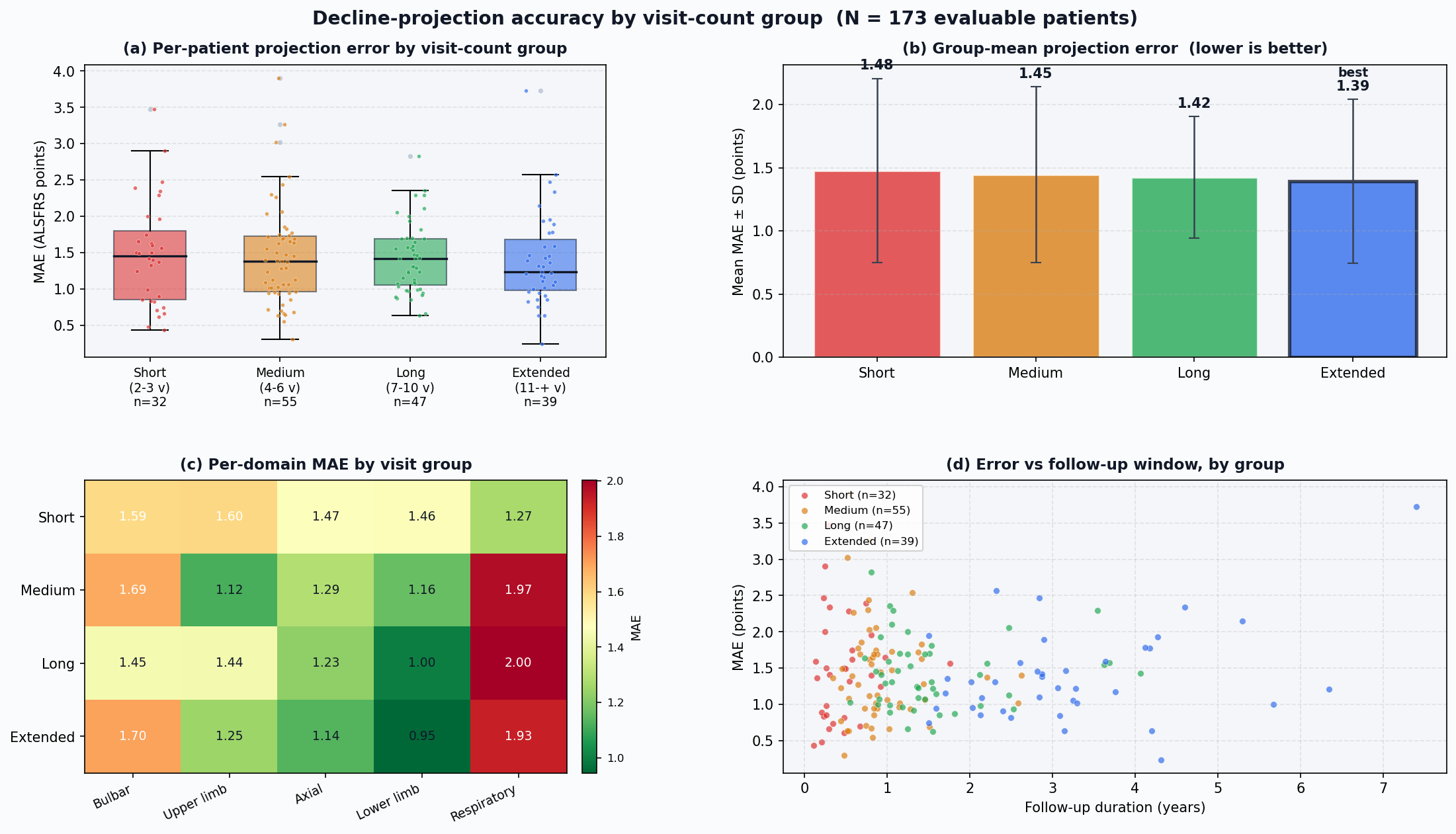}}
\caption{Decline-projection accuracy by visit-count group across
$N=173$ evaluable patients.
(a) Per-patient MAE distribution within each group;
(b) group-mean MAE with $\pm$1~SD error bars (best group highlighted);
(c) per-domain MAE heatmap (lower is better);
(d) per-patient MAE versus follow-up duration, coloured by group.
Groups defined by total visit count:
\emph{Short} (2--3 visits), \emph{Medium} (4--6), \emph{Long}
(7--10), \emph{Extended} ($\geq 11$).}
\label{fig:visit_groups}
\end{figure}

To quantify how follow-up density shapes projection accuracy, we
binned the 173 evaluable patients by their total visit count into four
groups chosen from the cohort distribution
(median = 6 visits, IQR 4--9):
\emph{Short} (2--3 visits, $n=32$),
\emph{Medium} (4--6, $n=55$),
\emph{Long} (7--10, $n=47$),
and \emph{Extended} ($\geq 11$, $n=39$).
For each patient, we computed the mean absolute error (MAE) and
root-mean-square error (RMSE) of the predicted ALSFRS-R trajectory
against held-out ground-truth scores
(Fig.~\ref{fig:visit_groups}).
MAE varied across the selected representative cases, whereas cohort-level mean MAE decreased with increasing follow-up density. ---
$1.48 \pm 0.73$ points for \emph{Short},
$1.45 \pm 0.70$ for \emph{Medium},
$1.42 \pm 0.48$ for \emph{Long},
and $1.39 \pm 0.65$ for \emph{Extended} ---
with a corresponding tightening of the per-patient error distribution
(IQR width $0.95 \rightarrow 0.77 \rightarrow 0.64 \rightarrow 0.70$).
The \emph{Extended} group, with a median follow-up duration of
1{,}057~days versus 119~days for \emph{Short}, achieved the lowest
mean error despite spanning the broadest temporal extrapolation,
indicating that additional input observations stabilise the latent
state estimate faster than longer prediction windows degrade it.
Per-domain analysis (Fig.~\ref{fig:visit_groups}c) showed that
respiratory and bulbar items remained the hardest to forecast across
all groups (MAE 1.27--2.00 and 1.45--1.70 respectively), whereas the
axial domain improved most as
follow-up grew (MAE 1.47 $\rightarrow$ 1.14).
Overall, these results confirm that the model preserves accuracy
across both rapid sub-year declines and multi-year slow-progressor
trajectories when sufficient input history is provided.

\subsection{Per-Domain Survival Performance}

Table~\ref{tab:results} presents the test-set evaluation for the ASTP model on the 28 held-out patients (194 landmark samples).
The model achieves C-index values ranging from 0.513 (Respiratory) to 0.598 (Lower limb), consistently above the random-chance baseline of 0.500.
Lower limb achieves the best discrimination, likely because gait decline in ALS follows a stereotyped pattern with strong temporal signal~\cite{kiernan2011als}.
Respiratory domain shows the weakest discrimination (C-index 0.513), which is consistent with the high baseline respiratory scores (mean 9.2 out of 12) and the known bimodal distribution of respiratory onset patterns in ALS.

\begin{table}[t]
\centering
\caption{Test Set Evaluation — ALS Domain Stage Transition Predictor (ASTP).
$n_\text{obs}$ = number of observed transitions; $n_\text{cens}$ = number of right-censored samples.
MAE and Median |error| computed over observed transitions only.}
\label{tab:results}
\begin{tabular}{lcccccc}
\toprule
Domain & $n_\text{obs}$ & $n_\text{cens}$ & C-index & MAE (d) & Med. |err.| (d) \\
\midrule
Bulbar       & 133 & 61  & 0.544 & 154.7 & 120.0 \\
Upper limb   & 130 & 64  & 0.543 & 160.8 & 120.0 \\
Axial  & 144 & 50  & 0.547 & 126.7 &  90.0 \\
Lower limb      & 124 & 70  & \textbf{0.598} & 169.4 & 120.0 \\
Respiratory  & 130 & 64  & 0.513 & 197.3 & 150.0 \\
\midrule
\textbf{Mean} & --- & --- & 0.549 & 161.8 & 120.0 \\
\bottomrule
\end{tabular}
\end{table}

\subsection{Per-Domain Survival NLL}

Table~\ref{tab:valloss} breaks down the best-epoch validation NLL per domain.
Bulbar and Respiratory domains achieve lower NLL (1.97 and 2.05 respectively), suggesting that these domains follow more predictable hazard patterns.
Upper limb and Axial show higher NLL ($\approx$2.5), consistent with their greater sensitivity to handedness-specific function loss and compensation strategies.

\begin{table}[t]
\centering
\caption{Per-Domain Validation NLL at Best Epoch (Epoch 19)}
\label{tab:valloss}
\begin{tabular}{lc}
\toprule
Domain & Val. NLL \\
\midrule
Bulbar       & 1.970 \\
Upper limb   & 2.502 \\
Axial  & 2.513 \\
Lower limb      & 2.244 \\
Respiratory  & 2.050 \\
\midrule
\textbf{Mean (total)} & \textbf{2.250} \\
\bottomrule

\vspace{-1cm}
\end{tabular}
\end{table}

\subsection{Mobility Decline Shows Strong Association with Earlier Healthcare Utilization}
\label{sec:cox_results}

Cox proportional hazards modeling (Table~\ref{tab:cox}) identified lower
limb functional decline as the strongest predictor of earlier care
utilization.
Impairments in walking (ALSFRS8: coef $= -0.30$, $p < 0.005$) and
climbing stairs (ALSFRS9: coef $= -0.40$, $p < 0.005$) were
significantly associated with increased hazard, indicating that reduced
mobility substantially accelerates the timing of assistive device use.
Follow-up time since diagnosis was also significantly associated with the
event ($p < 0.005$), reflecting the expected progression-dependent
increase in care needs over time.

Several additional functional measures demonstrated marginal associations
with healthcare utilization, including bulbar speech (ALSFRS1, $p = 0.08$) and
salivation (ALSFRS2, $p = 0.06$), as well as axial function (ALSFRS7,
$p = 0.05$).
Dressing and hygiene (ALSFRS5) showed a borderline positive association
($p = 0.05$), and sex exhibited a marginal effect ($p = 0.06$).
In contrast, age at visit and respiratory function measures
(ALSFRSR1--ALSFRSR3) were not significantly associated with time to healthcare
utilization.

\begin{table}[htbp]
\centering
\caption{Cox proportional hazards model for time-to-event prediction of healthcare
utilization. Negative coefficients indicate a lower hazard per one-unit increase in the covariate. Because higher ALSFRS-R scores represent better function, negative item coefficients imply that poorer function is associated with an earlier wheelchair event.
Lower limb items (ALSFRS8, ALSFRS9) and follow-up time are the strongest
predictors ($p < 0.005$).}
\label{tab:cox}
\begin{tabular}{lrrrrrr}
\toprule
\textbf{Covariate} & \textbf{coef} & \textbf{se} & \textbf{lo 95\%} & \textbf{hi 95\%} & \textbf{z} & \textbf{p} \\
\midrule
age\_at\_visit   &  0.00 & 0.01 & $-$0.02 &  0.01 & $-$0.58 & 0.56 \\
FU\_since\_dx    & $-$0.00 & 0.00 & $-$0.01 &  0.00 & $-$2.86 & $<$0.005 \\
sex              & $-$0.30 & 0.14 & $-$0.54 &  0.02 & $-$1.85 & 0.06 \\
ALSFRS1          & $-$0.10 & 0.05 & $-$0.19 &  0.01 & $-$1.74 & 0.08 \\
ALSFRS2          & $-$0.10 & 0.04 & $-$0.17 &  0.00 & $-$1.89 & 0.06 \\
ALSFRS3          &  0.00 & 0.06 & $-$0.11 &  0.11 & $-$0.05 & 0.96 \\
ALSFRS4          &  0.00 & 0.04 & $-$0.08 &  0.08 &  0.12 & 0.90 \\
ALSFRS5          &  0.30 & 0.16 &  0.00 &  0.60 &  1.93 & 0.05 \\
ALSFRS6          &  0.05 & 0.06 & $-$0.06 &  0.17 &  0.94 & 0.35 \\
ALSFRS7          & $-$0.10 & 0.05 & $-$0.19 &  0.00 & $-$1.97 & 0.05 \\
ALSFRS8          & $-$0.30 & 0.06 & $-$0.45 & $-$0.21 & $-$5.45 & $<$0.005 \\
ALSFRS9          & $-$0.40 & 0.05 & $-$0.53 & $-$0.34 & $-$8.87 & $<$0.005 \\
ALSFRSR1         & $-$0.00 & 0.04 & $-$0.11 &  0.06 & $-$0.52 & 0.60 \\
ALSFRSR2         &  0.00 & 0.04 & $-$0.07 &  0.07 & $-$0.02 & 0.99 \\
ALSFRSR3         &  0.06 & 0.07 & $-$0.07 &  0.19 &  0.91 & 0.37 \\
\bottomrule
\end{tabular}
\end{table}

\section{Discussion}
\label{sec:discussion}

In this study, we developed a time-to-event digital twin-inspired framework to
model ALS disease progression and predict clinically meaningful healthcare
utilization outcomes.
By integrating longitudinal ALSFRS-R trajectories, diagnostic metadata,
and assistive device records, our approach provides a unified
representation of disease dynamics that bridges functional decline and
real-world clinical events.

A key finding of this work is the dominant role of mobility decline,
particularly lower limb function, in driving earlier care utilization.
Both walking (ALSFRS8) and climbing stairs (ALSFRS9) emerged as the
strongest predictors of time to wheelchair access, significantly
outperforming other functional domains.
This observation is consistent with the known clinical progression of
ALS, where lower limb impairment often precedes more advanced disability
and triggers transitions to assistive care.

The longitudinal GAMM analysis further revealed that ALS progression is
characterized by strong, nonlinear temporal decline across all functional
domains, with coordinated but domain-specific deterioration patterns.
While all domains exhibited significant decline over time, the most rapid
changes were observed in motor-related functions, aligning with the
survival analysis findings.
In contrast, age showed minimal influence on disease trajectories, and
sex effects were limited to specific domains, suggesting that functional
decline in ALS is largely governed by disease-intrinsic mechanisms.

Another important contribution is the data-driven validation of ALSFRS-R
functional domains through correlation-based clustering.
The emergence of distinct bulbar, motor, and respiratory clusters
supports the biological coherence of these domains and justifies their
use in both longitudinal modeling and predictive frameworks.
The identification of subdomain structure within limb functions further
highlights the potential for more granular phenotyping of disease
progression.

Building on these findings, we implemented a Cox-based predictive model that enables individualized prediction of time to healthcare utilization.
By dynamically integrating patient-specific functional scores and
demographic features, the system generates interpretable survival curves
that reflect evolving risk profiles.

Despite these strengths, several limitations should be considered.
First, the analysis is based on observational data, which may be subject
to selection bias and unmeasured confounding.
Second, the Cox model assumes proportional hazards, which may not fully
capture complex, time-varying effects inherent in ALS progression.
Third, the framework complements rather than replaces individualized clinical assessment. Although trained on population-level data, it generates patient-specific risk estimates from each individual’s longitudinal history and updates them as new data become available. Predictions should therefore be interpreted alongside clinical judgment and patient preferences. Fourth, the predictive model prototype currently relies on structured clinical variables and does not incorporate additional modalities such as imaging,
genomics, or environmental factors.
Finally, external validation in independent cohorts is necessary to
assess the generalizability of the model.

Future work will focus on extending this framework to incorporate
multimodal data integration, including molecular and imaging biomarkers,
and developing an adaptive digital-twin-inspired predictive model that continuously updates predictions as new patient data become available. Future model extensions could incorporate time-varying coefficients and joint longitudinal-survival models to improve temporal flexibility \cite{zhong2021deep}.
Additionally, integrating mechanistic modeling with data-driven
approaches may further improve interpretability and clinical relevance.

\section{Conclusion}
\label{sec:conclusion}

We presented a time-to-event model for ALS that integrates longitudinal functional trajectories with survival modeling to enable individualized prediction of clinically meaningful healthcare utilization events.
By combining data-driven functional domain discovery, nonlinear longitudinal modeling, transformer-based stage transition prediction, and Cox proportional hazards–based risk estimation, the proposed framework captures both the temporal dynamics and clinical heterogeneity of ALS progression. The temporal modeling is an attention-based discrete-time survival model, not a neural sparse-regression method \cite{metayer2026data}. Rather than discovering governing equations, it predicts patient-specific time-to-event risks from irregular ALSFRS-R trajectories and generates domain-specific hazards for functional decline. The model avoids derivative estimation and predefined equation libraries, accommodates irregular follow-up, and provides dynamically updated, clinically interpretable survival probabilities within a digital-twin-inspired decision-support framework.
Our analyses consistently identified mobility decline—particularly impairments in walking and stair climbing—as the strongest predictors of earlier wheelchair utilization, highlighting lower limb dysfunction as a critical marker for proactive intervention and personalized care planning. 
The predictive model further demonstrates the feasibility of generating interpretable, patient-specific survival trajectories and dynamic risk predictions from routinely collected clinical data, supporting real-time clinical decision-making and longitudinal disease monitoring. 
Beyond ALS, this work could serve as a foundation for a scalable, extensible, and clinically actionable computational framework for precision neurodegenerative disease modeling, with future potential for multimodal integration of imaging, molecular, wearable, and environmental data to support adaptive, continuously updated digital twin systems for personalized medicine.

\section*{Acknowledgments}

The authors thank the participants, families, and clinical investigators
of the ALS Natural History Consortium who contributed data. This research is partially supported by the National Academies of Sciences, Engineering, and Medicine, grant number SCON-10001538 to Z.Y.

\section*{Accessibility}
The code is publicly available via: https://github.com/qilimk/als-digital-twin-app/

\bibliographystyle{IEEEtran}
\bibliography{references}

\end{document}